\documentclass{article}

\PassOptionsToPackage{numbers}{natbib}

     \usepackage[preprint]{neurips_2024}

\usepackage{amsmath}
\usepackage{diagbox}
\usepackage{caption}
\usepackage{amsfonts,amssymb}
\usepackage{subfigure}
\usepackage{graphicx}
\usepackage{bm}
\usepackage{multirow}
\usepackage[ruled,lined,commentsnumbered]{algorithm2e}
\usepackage{hyperref}
\usepackage{booktabs}
\usepackage{bbm}
\usepackage[inkscapelatex=false]{svg}
\usepackage{pifont}% http://ctan.org/pkg/pifont
\usepackage{stfloats}
\usepackage{multicol}
\usepackage{float}
\usepackage{array}
\newcommand{\PreserveBackslash}[1]{\let\temp=\\#1\let\\=\temp}
\newcolumntype{C}[1]{>{\PreserveBackslash\centering}p{#1}}
\newcolumntype{R}[1]{>{\PreserveBackslash\raggedleft}p{#1}}
\newcolumntype{L}[1]{>{\PreserveBackslash\raggedright}p{#1}}

\usepackage{mathrsfs}
\usepackage{wrapfig}

\usepackage[utf8]{inputenc} % allow utf-8 input
\usepackage[T1]{fontenc}    % use 8-bit T1 fonts
\usepackage{hyperref}       % hyperlinks
\usepackage{url}            % simple URL typesetting
\usepackage{booktabs}       % professional-quality tables
\usepackage{amsfonts}       % blackboard math symbols
\usepackage{nicefrac}       % compact symbols for 1/2, etc.
\usepackage{microtype}      % microtypography
\usepackage{xcolor}         % colors

\title{Agent-SiMT: Agent-assisted Simultaneous Machine Translation with Large Language Models}

\author{
  Shoutao Guo$^{1,3}$, Shaolei Zhang$^{1,3}$, Zhengrui Ma$^{1,3}$, Min Zhang$^{4}$, Yang Feng$^{1,2,3}$\thanks{ Corresponding author: Yang Feng.} \\
        \textsuperscript{\rm 1}{Key Laboratory of Intelligent Information Processing,} \\ Institute of Computing Technology, Chinese Academy of Sciences (ICT/CAS) \\
    { \textsuperscript{\rm 2} {Key Laboratory of AI Safety, Chinese Academy of Sciences}} \\
    { \textsuperscript{\rm 3} {University of Chinese Academy of Sciences, Beijing, China}} \\
{ \textsuperscript{\rm 4} {School of Future Science and Engineering, Soochow University}} \\
  {$\;\:$\texttt{\href{mailto:guoshoutao22z@ict.ac.cn}{guoshoutao22z@ict.ac.cn},\href{mailto:zhangshaolei20z@ict.ac.cn}{zhangshaolei20z@ict.ac.cn},\href{mailto:fengyang@ict.ac.cn}{fengyang@ict.ac.cn}}}
}

\begin{document}

\maketitle

\begin{abstract}
Simultaneous Machine Translation (SiMT) generates target translations while reading the source sentence. It relies on a policy to determine the optimal timing for reading sentences and generating translations. Existing SiMT methods generally adopt the traditional Transformer architecture, which concurrently determines the policy and generates translations. 
While they excel at determining policies, their translation performance is suboptimal. Conversely, Large Language Models (LLMs), trained on extensive corpora, possess superior generation capabilities, but it is difficult for them to acquire translation policy through the training methods of SiMT. Therefore, we introduce Agent-SiMT, a framework combining the strengths of LLMs and traditional SiMT methods. Agent-SiMT contains the policy-decision agent and the translation agent. The policy-decision agent is managed by a SiMT model, which determines the translation policy using partial source sentence and translation. The translation agent, leveraging an LLM, generates translation based on the partial source sentence. The two agents collaborate to accomplish SiMT. Experiments demonstrate that Agent-SiMT attains state-of-the-art performance.

\end{abstract}

\section{Introduction}
In recent years, simultaneous machine translation (SiMT) \citep{reinforcement, DBLP:conf/acl/MaHXZLZZHLLWW19, DBLP:conf/iclr/MaPCPG20, DualPath, zhao-etal-2023-adaptive} has been widely applied in real-time cross-language communication scenarios, such as international conferences and live broadcast. Unlike full-sentence translation \citep{DBLP:conf/nips/SutskeverVL14}, SiMT generates translations while reading the source sentence. To achieve the goals of low latency and high translation quality \citep{reinforcement, arivazhagan-etal-2019-monotonic}, SiMT relies on a policy to determine the optimal timing for reading source words and generating translations.

The existing SiMT methods \citep{zhang-etal-2022-learning, ma-etal-2023-non, DBLP:journals/corr/abs-2303-00257} generally adopt the traditional Transformer \citep{DBLP:conf/nips/VaswaniSPUJGKP17} to determine the policy and generate the translation concurrently. Based on the characteristics of the policy, these methods are typically divided into two categories: fixed policy and adaptive policy. For fixed policy, the model \citep{DBLP:conf/acl/MaHXZLZZHLLWW19, dalvi-etal-2018-incremental, DBLP:conf/emnlp/ZhangF21} produces a specific number of target words each time it reads a fixed number of source words. This heuristic approach ignores the reordering and length differences between the source and target sentences \citep{DBLP:conf/iclr/Gu0XLS18, guo-etal-2023-learning}. In contrast, the SiMT methods of adaptive policy \citep{DBLP:conf/iclr/MaPCPG20, liu-etal-2021-cross, ITST} can determine the appropriate timing for reading and generating words based on the translation context, achieving improved performance. However, it is a heavy burden for a Transformer model to carry out policy-decision and translation concurrently. Although adaptive SiMT methods can obtain superior policies \citep{DBLP:journals/corr/abs-2303-00257}, their translation capabilities are sub-optimal.

Large Language Models (LLMs) \citep{brown2020language, touvron2023llama, gemmateam2024gemma} have gained the most widespread attention in natural language processing, due to their exceptional abilities in understanding and generation. SiMT is expected to benefit from these strengths, yet there is limited exploration of LLMs for SiMT currently \citep{wang2024simultaneous, koshkin2024transllama}. This limitation can be attributed to the inherent complexities of SiMT, which pose significant challenges for the applications of LLMs. SiMT models of adaptive policy employ intricate training strategies to learn both policy-decision and translation generation concurrently. Despite possessing powerful generative abilities, LLMs struggle to acquire policy using traditional SiMT training methods due to their large-scale parameters and decoder-only architecture \citep{touvron2023llama}.

Under these grounds, we aim to combine the strengths of conventional SiMT models \citep{DBLP:journals/corr/abs-2303-00257} and LLMs \citep{touvron2023llama} through agent mechanism \citep{li2024agents}. We introduce Agent-SiMT, a novel framework designed to conduct SiMT with LLMs by employing a collaborative approach involving a policy-decision agent and a translation agent. The policy-decision agent employs a conventional SiMT model based on the encoder-decoder architecture \citep{DBLP:conf/nips/VaswaniSPUJGKP17} to make it compatible to utilize various existing SiMT models \citep{DBLP:journals/corr/abs-2303-00257, liu-etal-2021-cross}, while the translation agent adopts an LLM to take its advantage in understanding and generation. In addition to the use of the two agents that can collaboratively accomplish the problem of SiMT, Agent-SiMT also maintains a memory to deal with streaming input, which is used to store both input source words and generated translation. 

All components of Agent-SiMT work collaboratively as follows. Based on the source words and generated words in memory, the policy-decision agent first determines whether to read the source words into memory or to generate translation. Once the decision of generation is made, the translation agent will retrieve source words and translation in memory, and concatenate them as the prompt to proceed with the translation by predicting the next word. When the translation agent finishes generating the next target word, the newly generated word will be added to the translation in memory. Then a new working cycle begins with the policy-decision agent. Additionally, conventional SiMT models all employ the token-level policies \citep{ DBLP:journals/corr/abs-2303-00257}, and applying them directly to LLMs poses a vocabulary mismatch problem. 
To unify the policy representation across different models, 
we propose a novel word-level policy, which is derived from token-level policy and applied to LLMs.

The experiments demonstrate that, with a small amount of data fine-tuning open-source LLMs, our approach achieves significant improvement and attains state-of-the-art performance in SiMT. Further analysis shows the generalization of our approach and its usability in real-world scenarios.

\section{Background}
Our method utilizes the collaboration of the traditional Transformer-based SiMT model and an LLM to accomplish the SiMT task. Therefore, we provide a brief introduction to the SiMT task, Transformer-based SiMT model, and LLMs.

\textbf{Simultaneous Machine Translation} \; The SiMT model incrementally reads the source sentence $\mathbf{x}$ = $(x_1, ..., x_J)$ with length $J$ and generates translation $\mathbf{y}$ = $(y_1, ..., y_I)$ with length $I$ based on the translation policy. To describe the translation policy, we define $g_i$, which represents the number of input source words when generating the target word $y_i$. Then the policy for sentence pair ($\mathbf{x}, \mathbf{y}$) can be formulated as $\mathbf{g}$ = $(g_1, ..., g_I)$. During training, the SiMT model learns to generate target words using partial source words under the guidance of the policy $\mathbf{g}$ \citep{DBLP:conf/acl/MaHXZLZZHLLWW19}:
\begin{equation}
\mathcal{L}_{simt} = - \sum\limits_{i = 1}^{I} \log p(y_i \mid \mathbf{x}_{\leq g_i}, \mathbf{y}_{<i}, \theta_{simt}),
\end{equation}
where $\theta_{simt}$ represents the trained parameters of the SiMT model.

\textbf{Hidden Markov Transformer} \; Built upon the Transformer architecture \citep{DBLP:conf/nips/VaswaniSPUJGKP17}, Hidden Markov Transformer (HMT) \citep{DBLP:journals/corr/abs-2303-00257} is the currently most advanced SiMT method. By incorporating the Hidden Markov Model, it models the translation policy as the hidden events and the target sentence as observed events. Consequently, the HMT model implicitly learns to employ multiple policies for translation during training through the formulation:
\begin{equation}
\mathcal{L}_{hmt} = - \log \sum\limits_{\mathbf{g}} p ( \mathbf{y}\mid \mathbf{x},\mathbf{g},\theta_{hmt} ) \times p( \mathbf{g}, \theta_{hmt}),
\end{equation}
where $\theta_{hmt}$ represents the trained parameters of HMT.
During inference, HMT strategically selects the most suitable policy to guide the model in generating translations. In our approach, we exclusively harness its capability for policy-decision.

\textbf{Large Language Models} \;
Large Language Models (LLMs) mostly adopt the Decoder-only architecture \citep{brown2020language, touvron2023llama}. During training, LLMs are trained on large corpora using the maximum likelihood estimation objective. As a result, LLMs acquire powerful capabilities in understanding and generation. For the translation task, it concatenates the instruction, input source words $\mathbf{x}_{\leq j}$, and generated translation $\mathbf{y}_{<i}$ into a prompt as input to generate the next target word $y_i$:
\begin{equation}
 p ( y_i\mid \mathbf{x}_{\leq j},\mathbf{y}_{<i},\theta_{LLM} ),
\end{equation}
where $\theta_{LLM}$ signifies the parameters of the LLM. In our method, we leverage its translation capability.

\section{Method}
In this section, we introduce Agent-SiMT, which leverages agent collaboration to accomplish SiMT. We first outline the framework of Agent-SiMT and its working mechanism. To accomplish the vocabulary mismatch problem, we introduce a boundary-constrained word-level policy, which is derived from the token-level policy employed by conventional SiMT methods. Furthermore, we incorporate Supervised Fine-Tuning to enhance the translation capability of LLMs. The details will be introduced in the following sections.

\begin{figure*}[t]
    \centering
    \includegraphics[width=5.5in]{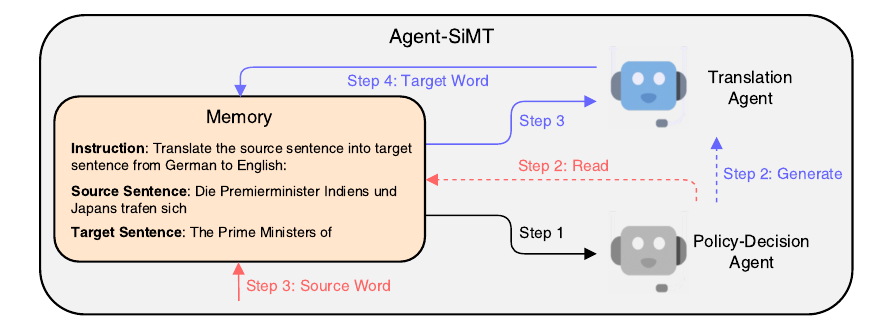}
    \caption{The framework of Agent-SiMT. The numbers in the diagram signify the execution sequence of Agent-SiMT. The red lines denote operations performed when the policy-decision agent determines to read source words. The blue lines indicate operations carried out when the decision for generation is made. The black line denotes the operation shared between both decision types.}
    \label{fig-model}
\end{figure*}

\subsection{Model Framework}
We introduce the model framework of Agent-SiMT. As illustrated in Figure \ref{fig-model}, Agent-SiMT mainly consists of three modules: memory, policy-decision agent, and translation agent. The memory module stores the instructions, input source words, and generated translation. The policy-decision agent is managed by the conventional SiMT model \cite{DBLP:journals/corr/abs-2303-00257}, which utilizes existing source words and generated target words to determine whether to read or generate a word. The translation agent employs an LLM \citep{touvron2023llama} to generate translation when the decision to generate words is made.

At each working cycle, the policy-decision agent retrieves the input source words and generated target words from memory to determine the action. If the policy-decision agent determines that the current source information is insufficient to generate the next word, it will instruct Agent-SiMT to read the next source word into the memory. If the decision to generate target words is made or the entire source sentence is read into the memory, the policy-decision agent will activate the translation agent. Subsequently, the translation agent retrieves instructions, existing source words, and generated translations from the memory module and then concatenates them as the prompt to generate the next target word. The newly generated word is then added to the memory module. Then a new working cycle begins with the policy-decision agent. Throughout the entire process, the two agents collaborate to accomplish the SiMT task.

In Agent-SiMT, existing SiMT models and LLMs can both serve as agents. In our main experiments, we choose \texttt{Llama2-7B-chat} \citep{touvron2023llama} as the translation agent and the current state-of-the-art SiMT model, the HMT model \citep{DBLP:journals/corr/abs-2303-00257}, as the policy-decision agent.

\subsection{Boundary-Constrained Word-level Policy}

\begin{wrapfigure}{r}{0.425\textwidth}
\begin{center}
\advance\leftskip+1mm
  \vspace{-0.2in} 
 \includegraphics[width=2.05in]{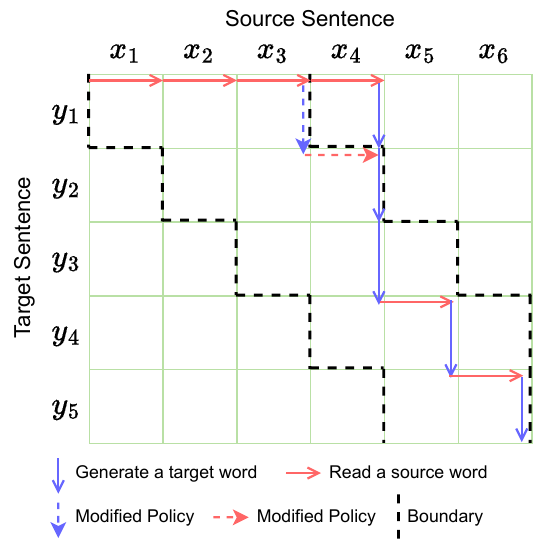}
  \caption{The illustration of incorporating boundary restrictions to word-level policy. The hyperparameters $B$ and $T$ in the figure are set to 1 and 3, respectively. In the absence of boundary restrictions, the word-level policy generates $y_1$ after reading $x_4$. However, our approach modifies it to generate $y_1$ upon reading $x_3$.
  }\label{boundary_constrain}
\vspace{-0.2in} 
\end{center}
\end{wrapfigure} 

We then introduce a boundary-constrained word-level translation policy. The conventional SiMT models \citep{DBLP:journals/corr/abs-2303-00257, liu-etal-2021-cross} yield a token-level policy. However, employing the token-level policy directly for translation poses challenges for LLMs, due to the vocabulary mismatch problem. Therefore, we introduce the word-level policy, which is derived from the token-level policy. Furthermore, to ensure that the performance is not affected by outlier policies \citep{DBLP:conf/iclr/MaPCPG20}, we incorporate boundary restrictions for the word-level policy. The methods for acquiring the word-level policy and implementing restrictions are introduced sequentially.

\textbf{Word-level Policy} \; Given the source sentence $\mathbf{x}$ = $(x_1, ..., x_J)$ and the target sentence $\mathbf{y}$ = $(y_1, ..., y_I)$, we define the corresponding source token sequence and target token sequence as $\mathbf{s}$ = $(s_1, ..., s_M)$ and $\mathbf{t}$ = $(t_1, ..., t_N)$, respectively. By employing the conventional SiMT model, we can obtain the token-level policy $\mathbf{h}$ = $(h_1, ..., h_N)$, where $h_n$ denotes the number of available source tokens when translating $t_n$. Subsequently, we introduce the method of transforming token-level policy into the word-level policy $\mathbf{g}$ = $(g_1, ..., g_I)$, where $g_i$ is the number of source words when translating $y_i$.

For the target token sequence $\mathbf{t}$, we iterate through each target token sequentially. When the target token $t_n$ is detected as the last token of a target word $y_i$, we obtain the number of source tokens $h_n$ used for translation at that moment. 
By assessing the number of complete source words formed by the first $h_n$ source tokens, we determine the number of complete source words as $u$ when translating $y_i$. Then $g_i$ is calculated as:
\begin{equation}
    g_i = \min \{u+1, J\},
\end{equation}
where $J$ is the length of the whole source sentence.
After traversing the entire target token sequence $\mathbf{t}$, we can obtain the word-level policy $\mathbf{g}$.

\textbf{Boundary Restrictions} \; Relying solely on the conventional SiMT model to determine the policy sometimes proves inadequate in preventing the emergence of outlier policies \citep{DBLP:journals/corr/abs-2303-00257}. These outlier policies may lead the model to generate translations by exploiting insufficient source information or result in excessively high latency \citep{DBLP:conf/iclr/MaPCPG20}. To address this problem, we impose restrictions on the obtained word-level policy $\mathbf{g}$ = $(g_1, ..., g_I)$. We then elaborate on the method of adding boundaries.

We define hyperparameters $B$ and $T$ to represent the minimum and maximum number of source words that can be considered when translating the first target word, respectively. The word-level policy with boundary restrictions can be obtained using Eq.(\ref{eq_1}) and Eq.(\ref{eq_2}) sequentially:
\begin{equation}
\label{eq_1}
    g^r_i = \min \{\max \{g_i, i-1+B\}, i-1+T\},
\end{equation}
\begin{equation}
\label{eq_2}
    g^r_i = \min \{g^r_i, J\},
\end{equation}
where $J$ is the length of the source sentence. Therefore, we obtain the word-level policy $\mathbf{g}^r$ = $(g^r_1, ..., g^r_I)$ with boundary restrictions. Figure \ref{boundary_constrain} provides an intuitive illustration of the word-level policy with boundary restrictions.

\subsection{Supervised Fine-Tuning}

After introducing the word-level policy, the modules of Agent-SiMT can collaborate to effectively carry out the SiMT task. To further improve the overall performance of Agent-SiMT, we fine-tune the LLM using Supervised Fine-Tuning (SFT), thereby enhancing the translation capability of LLM acting as the translation agent. As the translation agent, LLM typically utilizes partial source information to generate translations. Previous studies \citep{DBLP:conf/iclr/MaPCPG20, liu-etal-2021-cross, ITST} highlight that the SiMT model can demonstrate effective performance during inference only if it is trained to learn the ability to translate based on partial source sentence. However, there is no dedicated parallel corpus for the SiMT task, where the input is the source prefix and the output is the corresponding target prefix.

In our method, we fine-tune the LLM using a small amount of full-sentence parallel corpus. Notably, despite not being trained on SiMT corpora, our method exhibits a significant improvement over the conventional SiMT models. This underscores that our approach can effectively stimulate the LLM to excel in generating translations based on partial source information. During SFT, Agent-SiMT fine-tunes the LLM with LoRA \citep{hu2021lora}, and the details are introduced in Section \ref{Experiments}.
\section{Experiments}
\label{Experiments}
\subsection{Datasets}
We mainly validate our approach on three translation tasks.

\textbf{WMT15\footnote{\url{www.statmt.org/wmt15}} German$\rightarrow$English (De$\rightarrow$En)} This dataset comprises 4.5M parallel sentence pairs. During SFT, we randomly sample 100k sentence pairs for fine-tuning LLMs. Consistent with \citet{DBLP:conf/iclr/MaPCPG20}, we use newstest2013 as the validation set and newstest2015 as the test set.

\textbf{MuST-C English$\rightarrow$German (En$\rightarrow$De)} We conduct translation on text data \citep{di-gangi-etal-2019-must}. This dataset contains 230k samples. Similarly, we randomly sample 100k samples from the training set for fine-tuning LLMs. We use the dev set for validation and the tst-COMMON set for testing.

\textbf{WMT22\footnote{\url{www.statmt.org/wmt22}} Chinese$\rightarrow$English (Zn$\rightarrow$En)} Instead of using this dataset for fine-tuning LLMs, we evaluate the performance of our method on open-source translation LLMs \citep{jiao-etal-2023-parrot, zhang2023bayling} using the test set.

\subsection{System Settings}
Our experiments involve the following methods. Conventional SiMT methods employ Transformer architecture \citep{DBLP:conf/nips/VaswaniSPUJGKP17}. Our Agent-SiMT uses \texttt{Llama2-7B-chat}\footnote{\url{https://huggingface.co/meta-llama/Llama-2-7b-chat-hf}} as translation agent. 

\textbf{Full-sentence} is the conventional full-sentence machine translation model.

\textbf{Wait}-$k$ initially reads $k$ source words, and then generates a word and reads a word alternately \citep{DBLP:conf/acl/MaHXZLZZHLLWW19}.

\textbf{MMA} is the first Transformer-based SiMT model that performs the adaptive policy. It allows each head to independently determine its policy and integrates the outputs of multiple heads \citep{DBLP:conf/iclr/MaPCPG20}.

\textbf{ITST} models the SiMT task of translating source sentence to target sentence as an optimal transport problem and determines the policy by accumulating source information \citep{ITST}.

\textbf{HMT} is the state-of-the-art adaptive policy, utilizing Hidden Markov Model to model SiMT \citep{DBLP:journals/corr/abs-2303-00257}.

\textbf{Agent-Wait-$k$} adopts the Wait-$k$ policy as the policy-decision agent without fine-tuning LLMs.

\textbf{Agent-HMT} utilizes the HMT model as the policy-decision agent without fine-tuning LLMs.

\textbf{Agent-Wait-$k$+SFT} further fine-tunes the LLM upon Agent-Wait-$k$.

\textbf{Agent-HMT+SFT} fine-tunes the LLM based on Agent-HMT.

The SiMT methods based on Transformer are all adapted from Fairseq Library \citep{DBLP:conf/naacl/OttEBFGNGA19}. They apply Transformer-Base (6 layers, 8 heads) for the De$\rightarrow$En task and Transformer-Small (6 layers, 4 heads) for the En$\rightarrow$De task. The methods based on Agent-SiMT are all implemented using the Alpaca-LoRA Library\footnote{\url{https://github.com/tloen/alpaca-lora}}. During SFT, we fine-tune the LLM using LoRA \citep{hu2021lora}. For the adapters of LoRA, $r$ is set to 8, and $\alpha$ is set to 16. We set the learning rate to 0.0001 and batch size to 128. More training details and settings are shown in Appendix \ref{exper_detail}.

During inference, we apply greedy search to all methods. The methods based on Agent-SiMT adopt the word-level policy. We evaluate all methods with latency measured by Average Lagging (AL) \citep{DBLP:conf/acl/MaHXZLZZHLLWW19} and translation quality estimated by SacreBLEU \citep{post-2018-call}.

\subsection{Main Results}

\begin{figure*}[t]
\centering
\subfigure[The comparison of Agent-SiMT with previous Transformer-based SiMT methods on De$\rightarrow$En task.]{
\includegraphics[width=1.68in]{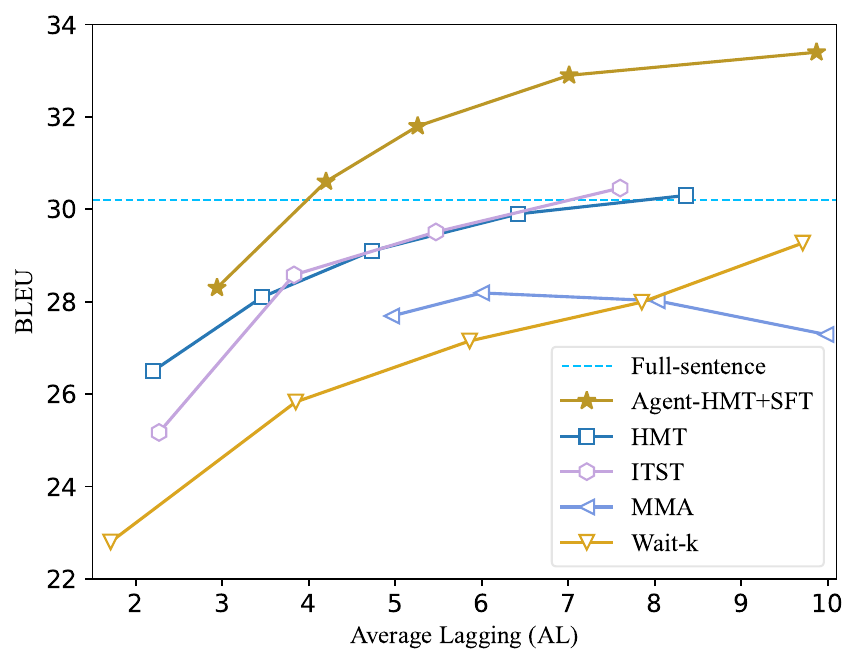}
}\hspace{0.1cm}
\subfigure[The comparison of HMT with the variants of Agent-SiMT on De$\rightarrow$En task.]{
\includegraphics[width=1.76in]{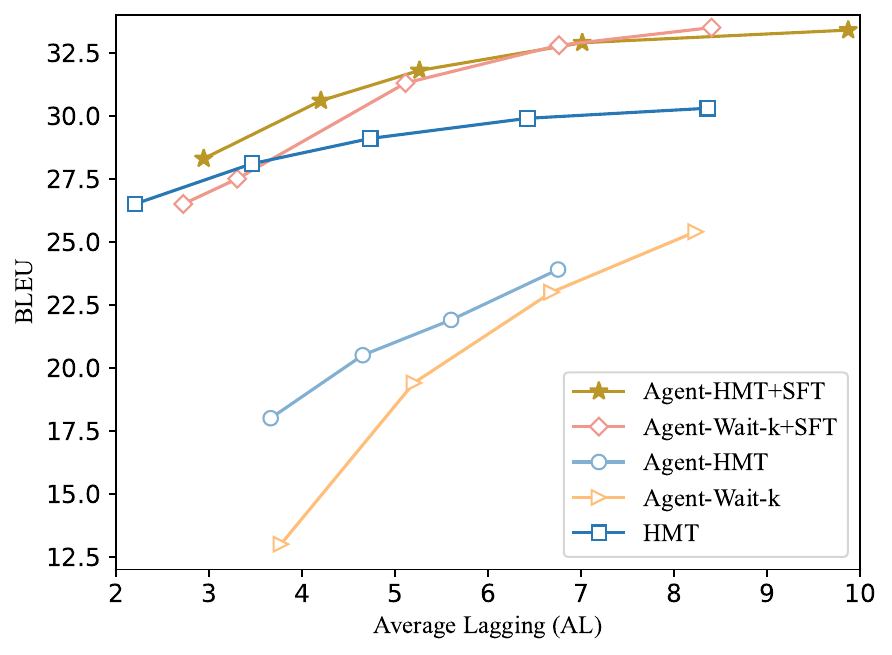}
}\hspace{0.1cm}
\subfigure[The comparison of HMT with the variants of Agent-SiMT on En$\rightarrow$De task.]{
\includegraphics[width=1.68in]{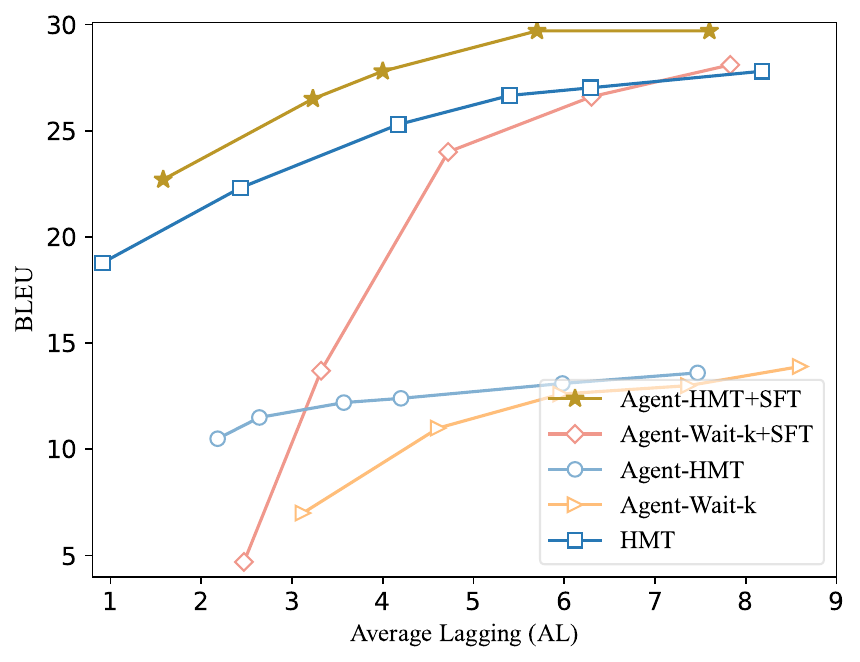}
}

\caption{Performance of different SiMT methods on De$\rightarrow$En and En$\rightarrow$De tasks.}
\label{main_res}
\end{figure*}

\begin{wrapfigure}{r}{0.35\textwidth}
\begin{center}
\advance\leftskip+1mm
  \vspace{-0.2in} 
 \includegraphics[width=1.75in]{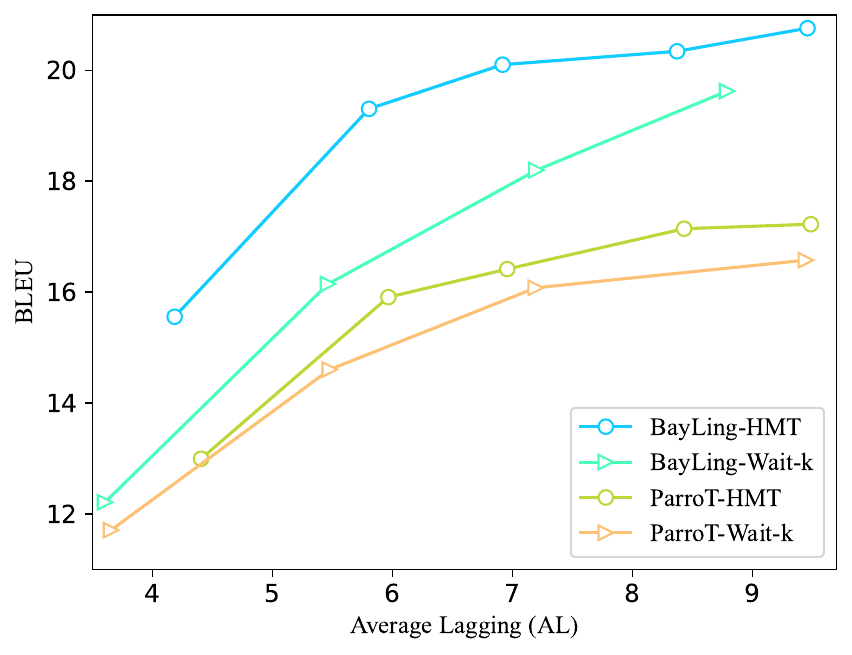}
  \caption{
Performance of our method on Zh$\rightarrow$En task when using different open-source translation LLMs as the translation agent.
  }\label{different_agent}
\vspace{-0.2in} 
\end{center}
\end{wrapfigure} 

We compare Agent-SiMT with previous SiMT methods built upon Transformer and explore different settings of Agent-SiMT. As shown in Figure \ref{main_res}, our method exhibits state-of-the-art performance.

Agent-HMT+SFT demonstrates a significant performance improvement over previous SiMT methods. Previous SiMT methods rely on a single model to concurrently manage policy-decision and translation \citep{DBLP:conf/iclr/MaPCPG20}. This exceeds the capacity of Transformer-based SiMT models, leading to degraded performance. In contrast, our approach decomposes the SiMT task into policy-decision and translation sub-tasks, assigning them to separate agents. Each agent is responsible for a specific sub-task and collaborates with the other agent to complete the SiMT, fully leveraging the strengths of each agent. Additionally, due to limitations in source information, the gap between Agent-HMT+SFT and traditional SiMT methods is relatively smaller at low latency. However, as latency increases, the divergence in translation capability becomes more pronounced.

Agent-HMT+SFT attains the best performance among all variants of Agent-SiMT. Compared to the Agent-Wait-$k$+SFT, Agent-HMT+SFT excels in policy-decision. For the policy-decision agent, HMT can dynamically adjust its policy \citep{DBLP:journals/corr/abs-2303-00257}, offering a more appropriate policy than the fixed policy of Wait-$k$ \citep{DBLP:conf/acl/MaHXZLZZHLLWW19}. This dynamic adjustment assists the translation agent in making well-informed decisions regarding translation timing, thereby achieving better performance. In comparison to Agent-HMT, the translation agent of Agent-HMT+SFT undergoes supervised fine-tuning, enhancing its translation capability. Consequently, enhancing the capabilities of each agent in Agent-HMT+SFT contributes to the overall performance.

To validate the generalization of our method, we also test its performance on the Zh$\rightarrow$En task using open-source translation LLMs \citep{jiao-etal-2023-parrot, zhang2023bayling}. We utilize ParroT\footnote{\url{https://github.com/wxjiao/ParroT}} and BayLing\footnote{\url{https://github.com/ictnlp/BayLing}} as translation agents and evaluate the performance with different policy-decision agents. The experimental results are shown in Figure \ref{different_agent}. Given a superior policy-decision agent, our method demonstrates strong SiMT performance with open-source translation LLMs.

\section{Analysis}
To enhance the comprehension of Agent-SiMT, we have undertaken various analyses. Unless otherwise specified, all analytical experiments are conducted on the De$\rightarrow$En task.

\subsection{Ablation Study}
\label{ablation}

\begin{wraptable}{r}{5.9cm}
\vspace{-0.1in} 
\caption{The performance of Agent-Wait-$k$+SFT when using different training data for SFT. `Full-Sentence' denotes full-sentence pairs. `SiMT' represents the data corresponding to the Wait-$5$ policy. The experiments are performed on De$\rightarrow$En task. }

\label{data_type}
\advance\leftskip+1mm
\small
\centering
\begin{tabular}{cccc}
\toprule
\textbf{SFT Data} & $k$ & \textbf{AL($\downarrow$)} & \textbf{BLEU($\uparrow$)}
\\ \cmidrule(lr){1-1} \cmidrule(lr){2-2}\cmidrule(lr){3-4}

Full-Sentence & 5 & \textbf{5.11} & \textbf{31.30} \\

SiMT & 5 & 5.43 &  1.00 \\

 \bottomrule     
\end{tabular}
\vspace{1.3in} 
\end{wraptable}
In the ablation experiments, we primarily investigate the influence of training data used for SFT on the performance of SiMT. Since SFT exclusively affects the translation agent, it essentially explores how the training data impacts the translation performance of the LLM.

Previous SiMT methods \citep{DBLP:conf/iclr/MaPCPG20, ITST} have found that SiMT models excel in SiMT during inference only when trained to generate translations based on partial source sentences. To validate this in Agent-SiMT, we construct the SiMT data corresponding to the Wait-$5$ policy to fine-tune the LLM and compare it with our method. As depicted in Table \ref{data_type}, full-sentence data proves more effective in stimulating the ability of LLMs to translate based on incomplete source information compared to SiMT data corresponding to the Wait-$k$ policy. We find that the translations generated by LLM deviate from the source sentences and lack fluency after using SiMT data for SFT. We attribute this issue to the fact that the SiMT data corresponding to the Wait-$k$ policy fails to consider the semantic equivalence between the source and target, resulting in the failure of SFT. The construction method for the SiMT data is detailed in the Appendix \ref{construct_method}.

\begin{wrapfigure}{r}{5.9cm}
\begin{center}
\advance\leftskip+1mm
  \vspace{-1.8in} 
 \includegraphics[width=1.75in]{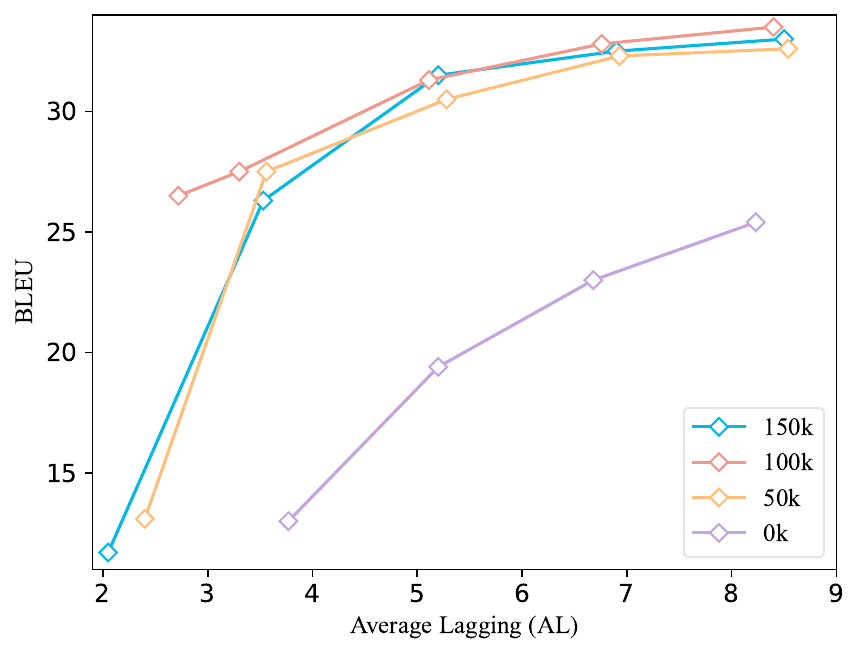}
  \caption{
The impact of different quantities of SFT training data on Agent-Wait-$k$+SFT. The experiments are on De$\rightarrow$En task.
  }\label{SFT_data_items}
\vspace{0.0in} 
\end{center}
\end{wrapfigure} 

Opting for fine-tuning the LLM with full-sentence data, we further investigate the impact of data quantity on the performance of Agent-SiMT. Figure \ref{SFT_data_items} illustrates that the performance of Agent-SiMT is significantly enhanced after SFT. Additionally, the impact of data quantity on SiMT performance appears to be relatively minimal. Therefore, we choose to fine-tune LLM with 100k samples, which demonstrates relatively better performance under all latency.

\subsection{Hallucination Rate}
\begin{wrapfigure}{r}{5.9cm}
\begin{center}
\advance\leftskip+1mm
  \vspace{-0.2in} 
 \includegraphics[width=1.98in]{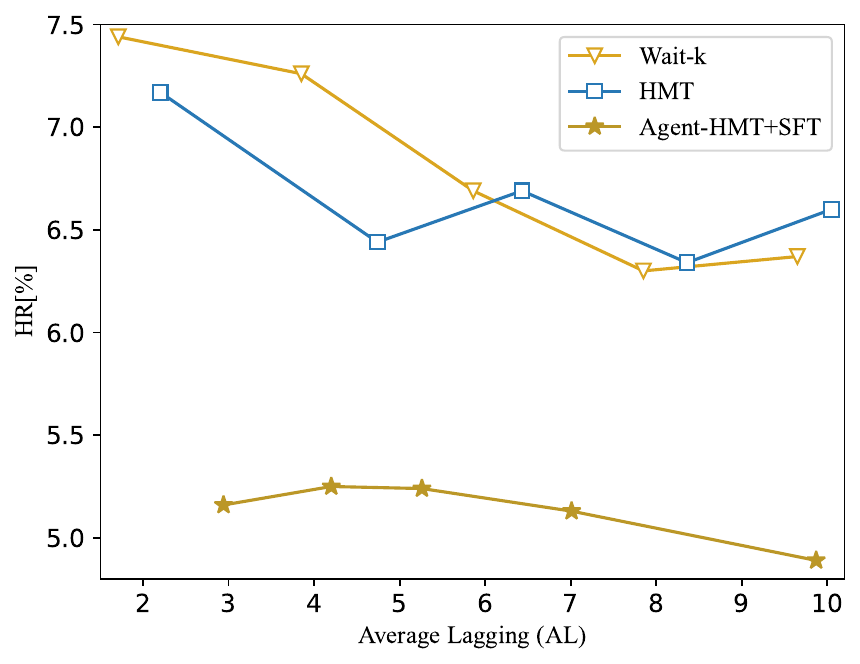}
  \caption{
The hallucination rate (HR) of different SiMT methods. The results are based on the De$\rightarrow$En task.
  }\label{hall}
 \vspace{-0.3in} 
\end{center}
\end{wrapfigure} 
In SiMT, hallucination typically refers to instances where the words in translation lack semantic counterparts in the source sentence. There are generally two main reasons for hallucinations in SiMT. On one hand, it stems from the insufficient translation capability of the SiMT model, resulting in the generation of unrelated words \citep{zhang2020futureguided}. On the other hand, SiMT models are compelled to utilize unrelated source information to predict target words under the guidance of a policy \citep{ma-etal-2023-non}. Therefore, the hallucinations in translation reflect a combination of translation and policy-decision capabilities.

To quantify the hallucination in translation, we introduce the hallucination rate (HR) \citep{DBLP:conf/emnlp/ChenZKM021}, which measures the proportion of hallucinated words in the translation. We employ \texttt{eflomal}\footnote{\url{https://github.com/robertostling/eflomal}} to get the alignments between the source sentence and translation, thereby calculating the HR for various SiMT methods. Figure \ref{hall} demonstrates that our method exhibits lower hallucination in translation under all latency. Compared to Wait-$k$, HMT achieves a lower HR due to its ability to adjust the policy based on the translation context. Our approach decomposes the SiMT into two sub-tasks and assigns them to different powerful agents. This maximizes the benefits of policy-decision and translation, achieving better trade-offs between latency and translation quality.

\subsection{Inference Speed}

The evaluation of SiMT tasks involves the assessment of latency and translation quality. However, the latency metric does not reflect the inference speed of machine \citep{DBLP:conf/acl/MaHXZLZZHLLWW19}. When incorporating the LLM into SiMT methods, a crucial question arises regarding whether the inference speed of LLMs could potentially act as an application bottleneck for the SiMT model. Therefore, it is important to compare the actual inference speed of our method with that of previous SiMT methods to demonstrate the practical usability of our approach.

\begin{wraptable}{r}{8.4cm}
\vspace{-0.1in}
\caption{Comparison of inference speed among different SiMT methods. `Speed' is measured by the number of words generated per second (i.e., word/s). The experiments are performed on the De$\rightarrow$En task. } 
\label{inference_speed}
\advance\leftskip+1mm
\small
\centering
\begin{tabular}{cccc}
\toprule
\textbf{Method} & \textbf{AL($\downarrow$)} & \textbf{BLEU($\uparrow$)} & \textbf{Speed$(\uparrow)$}                                       \\ \cmidrule(lr){1-1} \cmidrule(lr){2-3}\cmidrule(lr){4-4}

Wait-$k$ & 3.85 & 25.83 & \textbf{107.97} \\

HMT & 4.73 &  29.10 & 32.47 \\

Agent-HMT+SFT & \textbf{4.2} & \textbf{30.60} & 9.94 \\

 \bottomrule     
\end{tabular}

\end{wraptable}

We evaluate the inference speed of various SiMT methods using NVIDIA GeForce RTX 3090 and present the results in Table \ref{inference_speed}. The Wait-$k$ method exhibits the poorest performance but attains the highest inference speed due to its simplistic architecture. Although slower than the Wait-$k$, HMT demonstrates a significant improvement in performance. Our Agent-SiMT, incorporating HMT and LLMs, achieves a slower inference speed but optimal translation performance. 

\begin{wraptable}{r}{8.4cm}
\vspace{-0.1in}
\caption{Comparison of computation-aware latency and translation quality across different methods. AL-CA is measured in milliseconds, while COMET \citep{rei-etal-2020-comet} and BLEU are metrics used to evaluate machine translation. The experiments are performed on the En$\rightarrow$De task. } 
\label{caal}
\advance\leftskip+1mm
\small
\centering
\begin{tabular}{cccc}
\toprule
\textbf{Method} & \textbf{AL-CA($\downarrow$)} & \textbf{BLEU($\uparrow$)} & \textbf{COMET$(\uparrow)$}                                       \\ \midrule

\multirow{2}{*}{HMT} & 1677.92 & 26.63 & 74.72 \\

    & 2039.51 &  27.11 & 74.53 \\

\midrule

\multirow{2}{*}{Agent-HMT+SFT} & 2006.75 & 26.87 & 79.09 \\

    & 2549.60 &28.37 & 81.61 \\
 \bottomrule     
\end{tabular}

\end{wraptable}

From the analysis above, it can be observed that our method is slightly slower in inference speed compared to traditional Transformer-based SiMT methods. To demonstrate the practicality of our approach in real-world scenarios, we further compare the Computation-Aware Average Lagging (AL-CA) \citep{ma-etal-2020-simulmt, chang-etal-2022-anticipation} of different methods on the En$\rightarrow$De task. This computation-aware latency metric takes into account both machine inference delay and the lag time caused by waiting for source information to arrive. As shown in Table \ref{caal}, our method exhibits significantly better performance than HMT at roughly the same latency. 
Our method achieves good performance with approximately a 2000 ms delay in source sentences, making it suitable for practical applications.
Moreover, our method can benefit from acceleration on more advanced hardware, resulting in faster inference speeds.

\subsection{Performance at Different Levels of Difficulty}

The word reordering (i.e., non-monotonic alignment) between the source sentence and target sentence presents challenges for policy-decision and translation capabilities in SiMT. The translation difficulty varies across different sentence pairs, with long-distance non-monotonic alignments making translation notably more challenging. Addressing difficult sentence pairs reflects the comprehensive capabilities of SiMT method in policy-decision and translation. Consequently, we investigate the performance of SiMT models at different levels of translation difficulty. Based on the number and distance of non-monotonic alignment between the source sentence and ground-truth, we evenly divide the test set for the De$\rightarrow$En task into three levels: Easy, Medium, and Hard. Following the division, the Easy set primarily consists of monotonic alignments, while the Hard set includes sentences with at least 12 non-monotonic alignments \citep{DBLP:conf/emnlp/ZhangF21}.

\begin{wrapfigure}{t}{5.9cm}
\begin{center}
\advance\leftskip+1mm
  \vspace{-0.25in} 
 \includegraphics[width=1.7in]{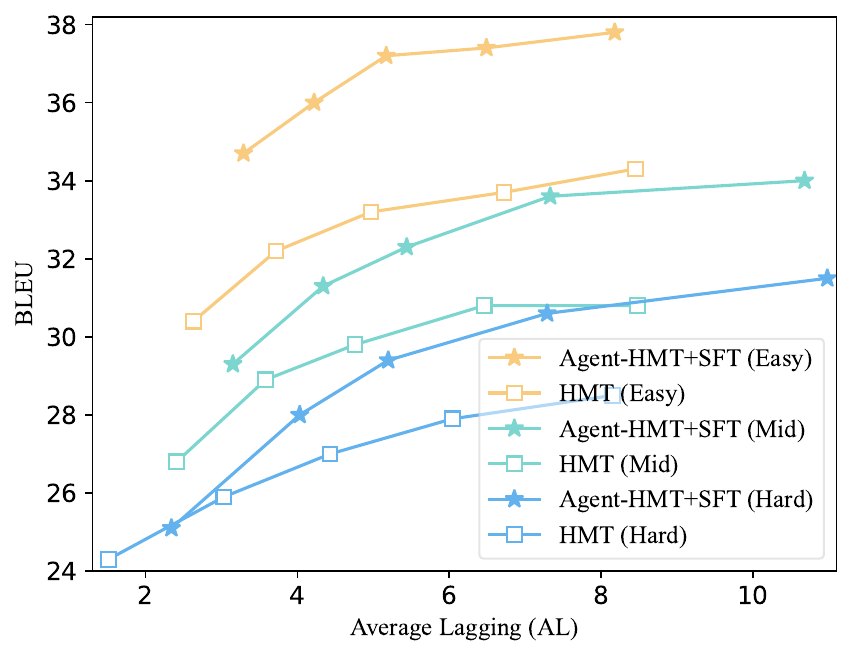}
  \caption{
Translation performance of SiMT methods at different levels of difficulty. The experiments are conducted on the De$\rightarrow$En dataset.
  }\label{difficult}
 \vspace{-0.4in} 
\end{center}
\end{wrapfigure} 

In Figure \ref{difficult}, we present the evaluation results of our method and HMT on datasets of three difficulty levels. Our method demonstrates substantial improvements over HMT at both easy and medium levels across all latency. However, at the difficult level, the performance of our method is relatively close to HMT at low latency. This is due to the extensive word reordering, which restricts the accessibility of source information for SiMT models at low latency. As the latency increases, the gap between our method and HMT also widens, underscoring the importance of task decomposition in our approach to prevent excessive burden on a single model.

\section{Related Work}
Simultaneous Machine Translation (SiMT) generates translations while reading the source sentence \citep{reinforcement}. This necessitates a translation policy to determine the number of source words when translating each target word. Consequently, SiMT entails the combination of policy-decision and translation. Depending on whether the policy is dynamically determined by the SiMT model, recent SiMT methods can be broadly categorized into two categories: fixed policy and adaptive policy.

\textbf{Fixed Policy} \; For fixed policy, \citet{DBLP:conf/acl/MaHXZLZZHLLWW19} first proposed the wait-$k$ policy, which initially reads $k$ source words, then alternate between generating a target word and reading a source word. Given the relatively singular nature of the fixed policy, most of the subsequent exploration employs the wait-$k$ policy or its variants \citep{dalvi-etal-2018-incremental, zeng-etal-2021-realtrans}, and focuses on improving training methods. Some methods \citep{DBLP:conf/emnlp/ChenZKM021, wang-etal-2023-better, guo-etal-2023-simultaneous} attempt to provide the model with references that maintain the same word order as the source sentences during training, thereby reducing the learning difficulty. Other methods mitigate the gap between SiMT and full-sentence translation by utilizing pseudo position information \citep{zhang-feng-2022-reducing}, curriculum learning \citep{10446517}, and distillation \citep{zhang2020futureguided} techniques. Additionally, ensemble learning is used to improve performance by combining the translation of multiple wait-$k$ policies \citep{DBLP:conf/emnlp/ZhangF21, elbayad2020efficient}.

\textbf{Adaptive Policy} \; In contrast to fixed policy, the methods
belonging to adaptive policy can assess the sufficiency of current source information to determine whether to read source words or generate translation, thereby achieving better tradeoffs between latency and translation quality. To learn superior policy, some methods \citep{DBLP:conf/iclr/MaPCPG20, DualPath, arivazhagan-etal-2019-monotonic, pmlr-v70-raffel17a} utilize expected attention to calculate the context vector, avoiding enumerating the exponential alignments between the source and target sentences. Subsequently, some methods \citep{liu-etal-2021-cross, tang-etal-2023-hybrid} utilize Transducer to handle SiMT, reducing the learning complexity by decoupling the temporal effects of the policy. Meanwhile, there are some methods, which acquire policy by comparing the prediction \citep{zhang-etal-2020-learning-adaptive} or distribution discrepancy \citep{zhao-etal-2023-adaptive} between SiMT and full-sentence translation. Additionally, some methods \citep{DBLP:conf/acl/ZhengLZMLH20, DBLP:conf/emnlp/GuoZF22} attempt to adapt the models of fixed policy into models performing adaptive policy. More recently, some works attempt to address SiMT problems using methods such as optimal transport theory \citep{ITST}, Hidden Markov Model \citep{DBLP:journals/corr/abs-2303-00257}, non-autoregressive models \citep{ma-etal-2023-non}, and achieve better SiMT performance.

Previous SiMT methods, while capable of achieving superior policies, exhibit sub-optimal translation performance. Conversely, LLMs demonstrate powerful generative capabilities but encounter challenges in mastering translation policies. Therefore, our Agent-SiMT combines the strengths of both conventional SiMT models and LLMs, employing agent collaboration to tackle SiMT problems.

\section{Conclusion}
In this paper, we introduce Agent-SiMT, a novel framework that harnesses existing Transformer-based SiMT methods and LLMs to collaboratively achieve SiMT. The experiments demonstrate that our method effectively leverages the strengths of different models, significantly surpassing previous SiMT methods. The analysis validates that our approach is practical for applications.

\section*{Limitations}
Our Agent-SiMT utilizes open-source LLMs as the translation agent and HMT as the policy-decision agent, resulting in significantly improved results. The experiments conducted in this paper demonstrate that enhancing either the translation or policy-decision capability can contribute to overall SiMT performance. Therefore, exploring more powerful LLMs and more efficient policy-decision models will further enhance SiMT performance.

\bibliographystyle{unsrtnat}
\bibliography{custom}

%%%%%%%%%%%%%%%%%%%%%%%%%%%%%%%%%%%%%%%%%%%%%%%%%%%%%%%%%%%%

\appendix

\section{Experiment Details}
\label{exper_detail}

\begin{figure*}[t]
    \centering
    \includegraphics[width=5in]{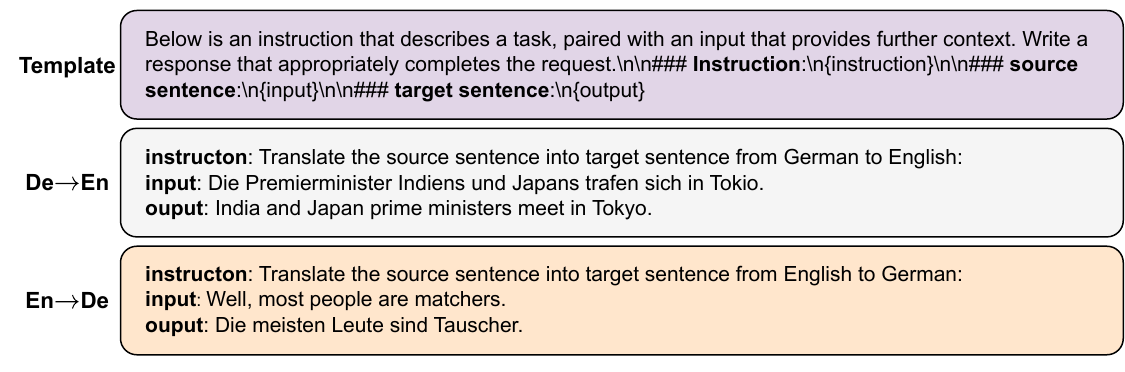}
    \caption{The prompt template for the LLM, along with examples for both De$\rightarrow$En and En$\rightarrow$De tasks.}
    \label{Instruct}
\end{figure*}

In this section, we primarily present the experimental setup for Agent-HMT+SFT. The settings of Agent-HMT+SFT are shown in Table \ref{tab:hyperparameter}. As shown in Figure \ref{Instruct}, we also provide the prompt template for the LLM, along with examples for both De$\rightarrow$En and En$\rightarrow$De tasks.

For the Zh$\rightarrow$En task, we do not fine-tune the LLMs. Instead, we evaluate the performance of our method using open-source translation LLMs \citep{jiao-etal-2023-parrot, zhang2023bayling} on the test set. The instruction templates for the open-source translation models ParroT \citep{jiao-etal-2023-parrot} and BayLing \citep{zhang2023bayling} are consistent with those used during training.

\begin{table*}[t]
\centering
\caption{Settings of Agent-HMT+SFT.}

\begin{tabular}{c|c|c|cc}
\toprule
\multicolumn{3}{c|}{\textbf{Hyperparameters}} & \textbf{WMT15 De$\rightarrow$En} & \textbf{MuST-C En$\rightarrow$De} \\
\bottomrule
\multirow{9}{*}{LLM} & Base\_model & Base\_model & \multicolumn{2}{c}{\texttt{Llama2-7B-chat}} \\
                    \cmidrule{2-5}
                     & \multirow{4}{*}{LoRA}  & lora\_r     & 8      & 8 \\
                     &                        & lora\_alpha & 16     & 16 \\
                     &                        & lora\_dropout & 0.05 & 0.05 \\
                     &                        & lora\_target\_modules & \multicolumn{2}{c}{q\_proj, k\_proj, v\_proj, o\_proj} \\
                     \cmidrule{2-5}
                     & \multirow{3}{*}{Training Details} & batch\_size & 128 & 128 \\
                     &                           & micro\_batch\_size & 4 & 4 \\
                     &                           & learning\_rate & 1e-4 & 1e-4 \\
\midrule
\multirow{19}{*}{HMT}  & \multirow{4}{*}{Encoder} & encoder\_layers & 6 & 6 \\
                     &                         & encoder\_embed\_dim & 512 & 512 \\
                    &                         & encoder\_ffn\_embed\_dim & 2048 & 1024 \\
                    &                           & encoder\_attention\_heads & 8 & 4 \\ 
                    \cmidrule{2-5}
                    & \multirow{4}{*}{Decoder} & decoder\_layers & 6 & 6 \\
                     &                         & decoder\_embed\_dim & 512 & 512 \\
                    &                         & decoder\_ffn\_embed\_dim & 2048 & 1024 \\
                    &                           & decoder\_attention\_heads & 8 & 4 \\
                    \cmidrule{2-5}
                    & \multirow{11}{*}{Training Details} & dropout   & 0.3    &0.3  \\
                    &                            &optimizer & adam  &adam   \\
                    &   &adam\_$\beta$          & (0.9, 0.98)   & (0.9, 0.98)     \\
                    & &clip\_norm        & 0             &0                \\
                    &&lr & 5e-4          & 5e-4                \\
                && lr\_scheduler   & inverse\_sqrt  & inverse\_sqrt     \\
                &&warmup\_updates       & 4000          & 4000            \\
&&warmup\_init\_lr     & 1e-7          & 1e-7                \\
&& weight\_decay       & 0.0        & 0.0        \\
&& label\_smoothing    & 0.1           & 0.1                \\
&&max\_tokens          &  8192$\times$4 & 8192$\times$4 \\
\bottomrule

\end{tabular}
\label{tab:hyperparameter}
\end{table*}

\section{Methods of Constructing SiMT Data}
\label{construct_method}
In subsection \ref{ablation}, we fine-tune the LLM using SiMT data corresponding to the Wait-$5$ policy. The methodology for constructing the SiMT data is elucidated in this section.

For a given sentence pair ($\mathbf{x}$, $\mathbf{y}$), with $\mathbf{x}$ having a length of $J$ and $\mathbf{y}$ having a length of $I$, we construct SiMT data corresponding to the Wait-$k$ policy, denoted as prefix pair. If the condition $k \geq J$ is satisfied, the prefix pair is ($\mathbf{x}$, $\mathbf{y}$). Otherwise, we randomly select $j$ from the interval $[k, J]$, where $J$ is the length of the source sentence. In this scenario, the source prefix is $\mathbf{x}_{\leq j}$. The length of the corresponding target prefix is calculated as:
\begin{equation}
    i = \min(j-k+1, I),
\end{equation}
where $I$ is the length of the target sentence. Consequently, the corresponding target prefix becomes $\mathbf{y}_{\leq i}$. At this point, we acquire the prefix pair ($\mathbf{x}_{\leq j}$, $\mathbf{y}_{\leq i}$).

\section{Numeric Results of Main Experiment}
In addition to the results shown in Figure \ref{main_res} and Figure \ref{different_agent}, we also provide the numeric results of the main experiment for better comparison. Table \ref{our_deen} demonstrates the results of HMT and Agent-SiMT on the WMT15 De$\rightarrow$En task. Table \ref{our_ende} gives the results of HMT and Agent-SiMT on the MuST-C En$\rightarrow$De task. Table \ref{our_zhen} gives the results of Agent-SiMT using open-source translation LLMs on the WMT22 Zh$\rightarrow$En task.

\begin{table}[ht]
\centering
\begin{tabular}{p{2cm}<{\centering} p{2cm}<{\centering} p{2cm}<{\centering}} 
\toprule[1.5pt]

\multicolumn{3}{c}{\textit{\textbf{HMT}}}     \\
\hline
    $L, K$      & AL  & SacreBLEU   \\
    2, 4      &2.20      &26.50   \\
    3, 6     &3.46      &28.10  \\
    5, 6      &4.73      &29.10  \\
    7, 6     &6.42      &29.90  \\
    9, 8     &8.36      &30.30 \\
\midrule[1pt]
\multicolumn{3}{c}{\textit{\textbf{Agent-Wait-$k$}}}     \\
\hline
    $k$      & AL  & SacreBLEU   \\
    1      &3.77      &13.00  \\
    3      &5.20      &19.40  \\
    5      &6.68      &23.00  \\
    7      &8.23      &25.40  \\
\midrule[1pt]
\multicolumn{3}{c}{\textit{\textbf{Agent-Wait-$k$+SFT}}}     \\
\hline
    $k$      & AL  & SacreBLEU   \\
    2      &2.72      &26.50  \\
    3      &3.30      &27.50  \\
    5      &5.11      &31.30  \\
    7      &6.76      &32.80  \\
    9      &8.40      &33.50  \\
\midrule[1pt]

\multicolumn{3}{c}{\textit{\textbf{Agent-HMT}}}     \\
\hline
    $L, K$    & AL  & SacreBLEU   \\
    2, 4   &2.20      &26.50   \\
    3, 6   &3.46      &28.10  \\
    5, 6    &4.73      &29.10  \\
    7, 6  &6.42      &29.90  \\
\midrule[1pt]

\multicolumn{3}{c}{\textit{\textbf{Agent-HMT+SFT}}}     \\
\hline
    $L, K$    & AL  & SacreBLEU   \\
    3, 6   &2.94      &28.30  \\
    5, 6    &4.20      &30.60  \\
    7, 6   &5.26      &31.80  \\
    9, 8   &7.01      &32.90   \\
    11, 8   &9.87      &33.40   \\
\bottomrule[1.5pt]

\end{tabular}
\vspace{0.1in} 
\caption{Numerical results of HMT and Agent-SiMT on De$\rightarrow$En task.}
\label{our_deen}
\end{table}

\begin{table}[tp]
\centering
\begin{tabular}{p{2cm}<{\centering} p{2cm}<{\centering} p{2cm}<{\centering}} 
\toprule[1.5pt]

\multicolumn{3}{c}{\textit{\textbf{HMT}}}     \\
\hline
    $L, K$   & AL  & SacreBLEU   \\
    1, 2   &0.92      &18.78   \\
    2, 2   &2.42      &22.31  \\
    4, 2    &4.16      &25.29  \\
    5, 4  &5.40      &26.66  \\
    6, 4    &6.29     &27.02  \\
    8, 6   &8.18      &27.80  \\
\midrule[1pt]
\multicolumn{3}{c}{\textit{\textbf{Agent-Wait-$k$}}}     \\
\hline
    $k$      & AL  & SacreBLEU   \\
    1      &3.12      &7.00  \\
    3      &4.61      &11.00  \\
    5      &5.96      &12.60  \\
    7      &7.37      &13.00  \\
    9      &8.60      &13.90  \\
\midrule[1pt]
\multicolumn{3}{c}{\textit{\textbf{Agent-Wait-$k$+SFT}}}     \\
\hline
    $k$      & AL  & SacreBLEU   \\
    1      &2.47      &4.70  \\
    3      &3.32      &13.70  \\
    5      &4.72      &24.00  \\
    7      &6.30      &26.60  \\
    9      &7.83      &28.10  \\
\midrule[1pt]

\multicolumn{3}{c}{\textit{\textbf{Agent-HMT}}}     \\
\hline
    $L, K$    & AL  & SacreBLEU   \\
    1, 2   &2.18      &10.50   \\
    2, 2   &2.64      &11.50  \\
    4, 2   &3.57      &12.20  \\
    5, 4  &4.20      &12.40  \\
    8, 6    &5.98      &13.10  \\
    9, 6   &7.47      &13.60  \\
\midrule[1pt]

\multicolumn{3}{c}{\textit{\textbf{Agent-HMT+SFT}}}     \\
\hline
    $L, K$    & AL  & SacreBLEU   \\
    4, 2    &1.58      &22.70  \\
    5, 4   &3.23      &26.50  \\
    6, 4   &4.00      &27.80  \\
    8, 6    &5.70     &29.70  \\
    9, 6   &7.60     &29.70  \\
\bottomrule[1.5pt]

\end{tabular}
\vspace{0.1in} 
\caption{Numerical results of HMT and Agent-SiMT on En$\rightarrow$De task.}
\label{our_ende}
\end{table}

\begin{table}[]
\centering
\begin{tabular}{p{2cm}<{\centering} p{2cm}<{\centering} p{2cm}<{\centering}} 
\toprule[1.5pt]

\multicolumn{3}{c}{\textit{\textbf{ParroT-Wait-$k$}}}     \\
\hline
    $k$      & AL  & SacreBLEU   \\
    3      &3.65      &11.71  \\
    5      &5.48      &14.61  \\
    7      &7.19      &16.08  \\
    9      &9.44      &16.58  \\
\midrule[1pt]
\multicolumn{3}{c}{\textit{\textbf{BayLing-Wait-$k$}}}     \\
\hline
    $k$      & AL  & SacreBLEU   \\
    3      &3.61      &12.21  \\
    5      &5.46      &16.15  \\
    7      &7.20      &18.20  \\
    9      &8.79      &19.63  \\
\midrule[1pt]

\multicolumn{3}{c}{\textit{\textbf{ParroT-HMT}}}     \\
\hline
    $L, K$    & AL  & SacreBLEU   \\
    1, 2   &4.41      &13.00  \\
    3, 6   &5.97      &15.91  \\
    5, 4   &6.96      &16.42  \\
    7, 6   &8.43      &17.15  \\
    9, 6   &9.49      &17.23  \\
\midrule[1pt]

\multicolumn{3}{c}{\textit{\textbf{BayLing-HMT}}}     \\
\hline
    $L, K$    & AL  & SacreBLEU   \\
    1, 2   &5.03      &15.56  \\
    3, 6   &6.37      &19.31  \\
    5, 4   &7.38      &20.10  \\
    7, 6   &8.77      &20.35  \\
    9, 6   &9.80      &20.76  \\
\bottomrule[1.5pt]

\end{tabular}
\vspace{0.1in} 
\caption{Numerical results of our method using open-source translation LLMs on Zh$\rightarrow$En task.}
\label{our_zhen}
\end{table}

\end{document}